\ifdefined\pdftexversion
  \pdfoutput=1
\fi
\documentclass[11pt]{article}
\ifdefined\pdfminorversion
\fi
\usepackage[final]{acl}

\usepackage[T1]{fontenc}
\usepackage[utf8]{inputenc}
\usepackage{times}
\usepackage{latexsym}
\usepackage{inconsolata}
\usepackage{microtype}
\usepackage{bm}

\usepackage{amsmath, amssymb, mathtools}
\AtBeginDocument{%
  \setlength{\abovedisplayskip}{4pt plus 2pt minus 2pt}%
  \setlength{\belowdisplayskip}{4pt plus 2pt minus 2pt}%
  \setlength{\textfloatsep}{10pt plus 2pt minus 2pt}%
  \setlength{\dbltextfloatsep}{8pt plus 2pt minus 2pt}%
}

\usepackage{graphicx}
\usepackage{caption}
\usepackage{subcaption}
\usepackage{booktabs}
\usepackage{makecell}
\usepackage{multirow}
\usepackage{balance}
\newsavebox{\tablebox}
\newcommand{\fitTable}[1]{%
  \sbox{\tablebox}{#1}%
  \ifdim\wd\tablebox>\linewidth
    \resizebox{\linewidth}{!}{\usebox{\tablebox}}%
  \else\usebox{\tablebox}\fi
}
\newcommand{\resultci}[3]{%
  {\normalsize #1}$_{\scriptstyle [#2,\,#3]}$%
}
\usepackage{float}

\usepackage{tikz}
\usetikzlibrary{arrows.meta,positioning}

\newcommand{\acronym}{{\sc{CRL-MMNAR}}}

\usepackage[ruled,vlined,linesnumbered,algo2e]{algorithm2e}
\usepackage{algorithm,algorithmicx,algpseudocode}

\usepackage{comment}
\usepackage{enumitem}

\usepackage[most]{tcolorbox}
\tcbset{
  myalgo/.style={
    colback=gray!5!white,
    colframe=black!75!black,
    fonttitle=\bfseries,
    title=Algorithm 1: MMNAR-Aware Modality Fusion,
    boxrule=0.5pt,
    arc=1mm,
    top=1mm, bottom=1mm, left=1mm, right=1mm,
    sharp corners,
    enhanced,
    breakable
  }
}

\title{Causal Representation Learning from Multimodal Clinical Records under Non-Random Modality Missingness}

\author{
  Zihan Liang\thanks{Equal contribution. Code and reproducibility materials are available at \url{https://github.com/CausalMLResearch/CRL-MMNAR}.} \\
  Emory University \\
  \texttt{zihan.liang@emory.edu}
  \And
  Ziwen Pan\footnotemark[1] \\
  Emory University \\
  \texttt{ziwen.pan@emory.edu}
  \And
  Ruoxuan Xiong \\
  Emory University \\
  \texttt{ruoxuan.xiong@emory.edu}
}

\begin{document}
\raggedbottom
\maketitle
\begin{abstract} 
Clinical notes contain rich patient information, such as diagnoses or medications, making them valuable for \textit{patient representation learning}. Recent advances in large language models have further improved the ability to extract meaningful representations from clinical texts. However, clinical notes are often missing. For example, in our MIMIC-IV cohort, $35.15\%$ of patients have no eligible discharge summaries. In such cases, representations can be learned from other modalities such as structured data, chest X-rays, or radiology reports. Yet the availability of these modalities is influenced by clinical decision-making and varies across patients, potentially producing modality missing-not-at-random (\textit{MMNAR}) patterns. We propose a \textit{causal representation learning} framework that leverages observed data and informative missingness in multimodal clinical records. It consists of: (1) an MMNAR-aware modality fusion component that integrates structured data, imaging, and text while conditioning on missingness patterns to capture patient health and clinician-driven assignment; (2) a modality reconstruction component with latent alignment to preserve modality information in the learned representation; and (3) a multitask outcome prediction model with a rectifier that corrects for residual bias from specific modality observation patterns. 
Comprehensive evaluations across MIMIC-IV and eICU show consistent gains over the strongest baselines, achieving relative AUC improvements of up to $\it{4.3\%}$ for hospital readmission and $\it{2.7\%}$ for ICU admission.

\end{abstract}

\section{Introduction}
\label{intro}

Language plays a central role in clinical communication. Learning patient representations from clinical notes has become an important focus in clinical NLP. These unstructured texts--written by clinicians to document observations, diagnoses, and decisions--encode rich contextual information that complements structured patient data. Since the introduction of contextualized language models like BERT \citep{devlin2019bert}, the field has advanced rapidly with medical-domain adaptations, such as ClinicalBERT \cite{alsentzer2019publicly,huang2019clinicalbert}. More recently, large language models (LLMs) fine-tuned or adapted to clinical tasks have shown promise in medical reasoning, outcome prediction, and clinical decision support \cite{yang2022large,singhal2023large,agrawal-etal-2022-large}. 

Yet clinical text is often missing in real-world settings. In our cohort from MIMIC-IV, a large publicly available collection of de-identified electronic health records (EHR) \cite{johnson2024mimic}, $35.15\%$ of patients lack eligible discharge summaries. In such cases, other EHR modalities such as structured data, chest X-rays (CXR), and radiology reports may still be available and can be leveraged to learn patient representations.

\begin{figure}[t!]
    \centering
    \includegraphics[width=1\linewidth]{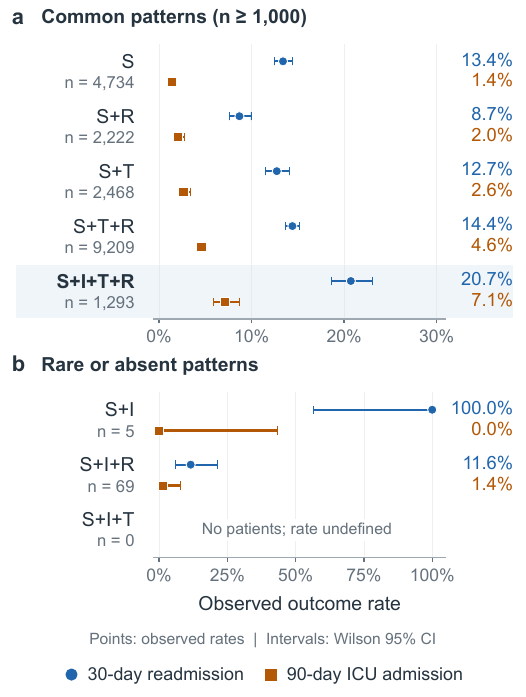}
    \caption{\textbf{Clinical outcome occurance rates vary across modality patterns.} S = structured EHR data; 
    I = chest X-ray images; 
    T =  discharge summaries; 
    R = radiology report. Points show observed proportions and horizontal intervals show Wilson 95\% confidence intervals in the $20{,}000$-patient cohort. Common patterns ($n\geq1{,}000$) and rare or absent patterns use different horizontal scales. S is observed for all patients. S+I+T has no patients, so its rates are undefined. Counts are provided in Table~\ref{tab:pattern_outcomes}.
    }
    \label{fig:modality_outcome}
\end{figure}

Crucially, the availability of these modalities is \textbf{not random}. It reflects physician decision-making, institutional protocols, patient conditions, and linked-dataset coverage. For example, clinical notes and radiology reports are more likely to be recorded for patients with more severe conditions or complex diagnostic needs. As shown in Figure~\ref{fig:modality_outcome} and Table~\ref{tab:pattern_outcomes}, outcome occurance rates vary across modality patterns. Patients with all four modalities have a $30$-day readmission rate of $20.7\%$, compared with $13.3\%$ among the remaining patients; the corresponding $90$-day post-discharge ICU admission rates are $7.1\%$ and $3.2\%$. These associations support our motivation for modeling modality missing-not-at-random (MMNAR) patterns, where data availability carries information about patient state and correlates with outcomes.

In this paper, we propose \acronym{} (\textit{Causal Representation Learning under MMNAR}), a novel framework that explicitly leverages both observed data and informative missingness in multimodal clinical records. Our framework comprises three components organized into two functional stages. Together, they improve patient representations when clinical notes are available and support representation learning when text is missing. 

The first stage consists of two complementary components for patient representation learning. The first component is \textbf{MMNAR-aware modality fusion}, which integrates representations of structured data, imaging, and text, while explicitly conditioning on modality missingness patterns. It serves three objectives: (i) increasing the estimation precision of latent patient representation by combining signals shared across modalities; (ii) preserving modality-specific information; and (iii) retaining signals associated with clinical decision-making, such as decisions to order labs or imaging. By combining multimodal content with the observation pattern, the fused representation captures health-related information alongside signals about clinical data collection.

The second component is \textbf{modality reconstruction with latent alignment}. Its purpose is to encourage the fused representation to preserve information from each modality and recover masked modality representations. A modality-specific predictor reconstructs the latent representation of a randomly masked modality from the remaining inputs, and a normalized alignment loss brings the reconstruction close to the target representation of the observed modality. This objective encourages cross-modal consistency and generalization across missingness patterns.

The second stage is \textbf{multitask outcome prediction}, where the learned patient representations are applied to downstream tasks such as $30$-day readmission, post-discharge ICU admission, and mortality. In MIMIC-IV, all three tasks use a common retrospective information window referenced to index hospital discharge, with mortality defined as $90$-day post-discharge all-cause mortality. The patient representation from Stage 1 serves as a shared backbone for related outcome tasks, improving statistical efficiency by pooling information from common features of patient health. On top of this backbone, task-specific heads capture heterogeneity unique to each clinical outcome.

We treat the modality observation pattern as a treatment variable that summarizes clinician-assigned data collection and associated care decisions, such as ordering additional tests or initiating treatment. Beyond informing the patient representation, this assignment can affect outcomes. We therefore introduce a \textbf{rectifier mechanism} that models observation-pattern-specific treatment effects through an additive outcome decomposition. Inspired by semiparametric debiasing methods \citep{robins1994estimation,robins1995semiparametric}, the rectifier estimates the treatment-effect component not captured by the base predictor using held-out residuals.

We validate our approach with extensive experiments on two large-scale clinical datasets, MIMIC-IV and eICU \cite{pollard_eicu_2018}. 
Our method consistently outperforms $13$ baselines. On MIMIC-IV, it achieves relative AUC gains of $\mathbf{+4.3\%}$ for readmission (from $0.741$ to $0.773$),
$\mathbf{+2.7\%}$ for ICU admission (from $0.845$ to $0.868$),
and $\mathbf{+0.7\%}$ for $90$-day post-discharge mortality (from $0.950$ to $0.957$).
On eICU, our method achieves a $\mathbf{+2.2\%}$ relative gain for readmission (from $0.801$ to $0.819$)
and $\mathbf{+1.9\%}$ for mortality (from $0.928$ to $0.946$). Subgroup analyses characterize performance under different modality configurations, while ablation studies on both datasets show incremental gains from all three components.

Although we focus on healthcare as the primary application, we note that the central idea extends more broadly. Modality observation patterns often contain meaningful signals about the processes that generate the data and influence outcomes.

\section{Problem Formulation}\label{sec:problem}

Let $\mathcal{M}$ be the set of all available modalities. For each patient $i$, let $\mathcal{M}_i \subseteq \mathcal{M}$ denote the subset of modalities actually observed. For any modality $m \in \mathcal{M}$, let $x_i^{(m)}$ denote the corresponding raw input for patient $i$. We define $\bm{x}_i = \{x^{(m)}_i\}_{m \in \mathcal{M}}$ as the collection of all modality inputs (whether observed or not), and $\bm{x}^{\text{obs}}_i = \{x_i^{(m)}\}_{m \in \mathcal{M}_i}$ as the subset of observed modalities for patient $i$.

To encode the modality observation pattern, we define a binary vector $\bm\delta_i = [\delta_i^{(m)}]_{m \in \mathcal{M}}$. This pattern reflects clinician decisions, such as whether to order imaging or write detailed notes, as well as source coverage and the observation protocol. For each modality $m \in \mathcal{M}$, we set $\delta_i^{(m)} = 1$ if it is observed for patient $i$, and $0$ otherwise. For example, in MIMIC-IV, we have $\mathcal{M} = \{\text{S}, \text{I}, \text{T}, \text{R}\}$, where ``S''  represents structured EHR data, ``I'' chest X-ray images, ``T'' discharge summaries, and ``R'' radiology reports. In this case, $\bm\delta_i$ is a four-dimensional binary vector describing the observed modality configuration. Modality availability is determined from inputs eligible under the index-discharge observation protocol (Appendix~\ref{appendix:data_preprocessing}).

For each patient $i$, we consider multiple clinical outcomes of interest, such as 30-day hospital readmission, post-discharge ICU admission within 90 days, and mortality. The mortality endpoint is 90-day post-discharge all-cause mortality in MIMIC-IV and in-hospital mortality in eICU. Let $\mathcal{T}$ denote the outcome tasks evaluated for a given dataset. For each $t \in \mathcal{T}$, let $y_{i,t}\in\{0,1\}$ denote the outcome for patient $i$ corresponding to task $t$ and let $\bm{y}_i = [y_{i,t}]_{t \in \mathcal{T}}$ collect all outcomes.

Our goal is to build a predictive model that uses both the observed modalities and the observation pattern to estimate outcomes as accurately as possible. Formally, we define the outcome model as
\begin{equation}\label{eqn:outcome-model}
    y_{i,t} = f_{\bm{\theta}_t}\Big(\bm{x}^{\text{obs}}_i, \bm\delta_i\Big) + \varepsilon_{i,t} \, ,
\end{equation}
where $f_{\bm{\theta}_t}$ is the prediction function parameterized by $\bm{\theta}_t$, and $\varepsilon_{i,t}$ denotes random noise. 

\begin{figure}[t]
\centering
\begin{tikzpicture}[
    >=Latex,
    node distance=18mm,
    every node/.style={circle,draw,minimum size=8mm,inner sep=0pt},
    bend angle=18
]
\node (h) {$\bm u_i$};
\node (x) [above left=4mm and 16mm of h] {$\bm{x}_i$};
\node (y) [right=16mm of h] {$\bm{y}_i$};
\node (d) [below left=4mm and 16mm of h] {$\bm{\delta}_i$};

\draw[->] (h) -- (x);          
\draw[->] (h) -- (y);          
\draw[->] (h) -- (d);          
\draw[->] (x) -- (d);
\draw[->, bend right=18] (d) to (y); 
\end{tikzpicture}
\caption{Causal motivation: The latent health state $\bm u_i$ influences modality contents $\bm x_i$, the observation pattern $\bm\delta_i$, and outcomes $\bm y_i$. The arrow from $\bm\delta_i$ to $\bm y_i$ summarizes possible effects of care decisions associated with modality collection. The learned representation $\bm h_i$ summarizes observed records and their availability; it is distinct from the unobserved state $\bm u_i$. }
\label{fig:causal-diagram}
\end{figure}

\section{Method}\label{methods}

Our \acronym{} method learns the outcome model \eqref{eqn:outcome-model} under the causal diagram in Figure~\ref{fig:causal-diagram}. 
The diagram posits a latent patient health state $\bm u_i$ that influences modality contents $\bm x_i$, observation patterns $\bm\delta_i$, and outcomes $\bm y_i$. We treat $\bm\delta_i$ as a treatment variable summarizing clinician-assigned modality collection and associated care decisions, with a direct effect on outcomes in addition to the influence of $\bm u_i$. We learn a representation $\bm h_i$ of the information about this state and the observation process available in the records. \acronym{} comprises three components organized into representation learning and outcome prediction.

In the first stage, we learn a patient representation $\bm{h}_i$ that captures both observed modalities and the observation pattern:
\begin{equation}\label{eqn:represenation-learning}
    \bm{h}_i = r_{\bm{\eta}} \Big(\bm{x}^{\text{obs}}_i, \bm{\delta}_i\Big) \, ,
\end{equation}
where $r_{\bm{\eta}}$ shares representation parameters across outcomes. In MIMIC-IV, the three tasks use a common retrospective information window referenced to discharge. This stage corresponds to Sections \ref{method:preprocess}-\ref{method:rep-learning}, covering preprocessing of raw data together with the two core components: modality fusion and modality reconstruction.

In the second stage, we adopt a multitask outcome prediction framework. For each outcome task $t \in \mathcal{T}$, we model 
\begin{equation}\label{eqn:multitask-outcome-prediction}
    y_{i,t} = g_{\bm{\psi}_t}\big(\bm{h}_i, \bm{\delta}_i\big) + \varepsilon_{i,t} \, ,
\end{equation}
where $g_{\bm{\psi}_t}$ is a task-specific predictor with parameters $\bm{\psi}_t$. Here, $\bm\delta_i$ is explicitly included as the treatment assignment because its effect on the outcome may not be fully captured by the shared representation. A rectifier mechanism (Section~\ref{method:multitask_outcome}) estimates the observation-pattern-specific treatment-effect component through held-out residuals and incorporates it into outcome prediction.

Finally, joint training integrates modality fusion, modality reconstruction, and outcome prediction, as summarized in Section \ref{method:training}. For each outcome task $t$, the parameter set is $\bm{\theta}_t = (\bm{\eta}, \bm{\psi}_t)$. The two stages give $f_{\bm{\theta}_t}(\bm{x}^{\mathrm{obs}}_i,\bm\delta_i)=g_{\bm{\psi}_t}(\bm h_i,\bm\delta_i)$, with $\bm h_i=r_{\bm\eta}(\bm{x}^{\mathrm{obs}}_i,\bm\delta_i)$. A complete algorithmic summary of the framework is provided in Appendix~\ref{appendix:algorithm}.

\begin{figure*}[ht]
  \centering
  \includegraphics[width=0.77\textwidth]{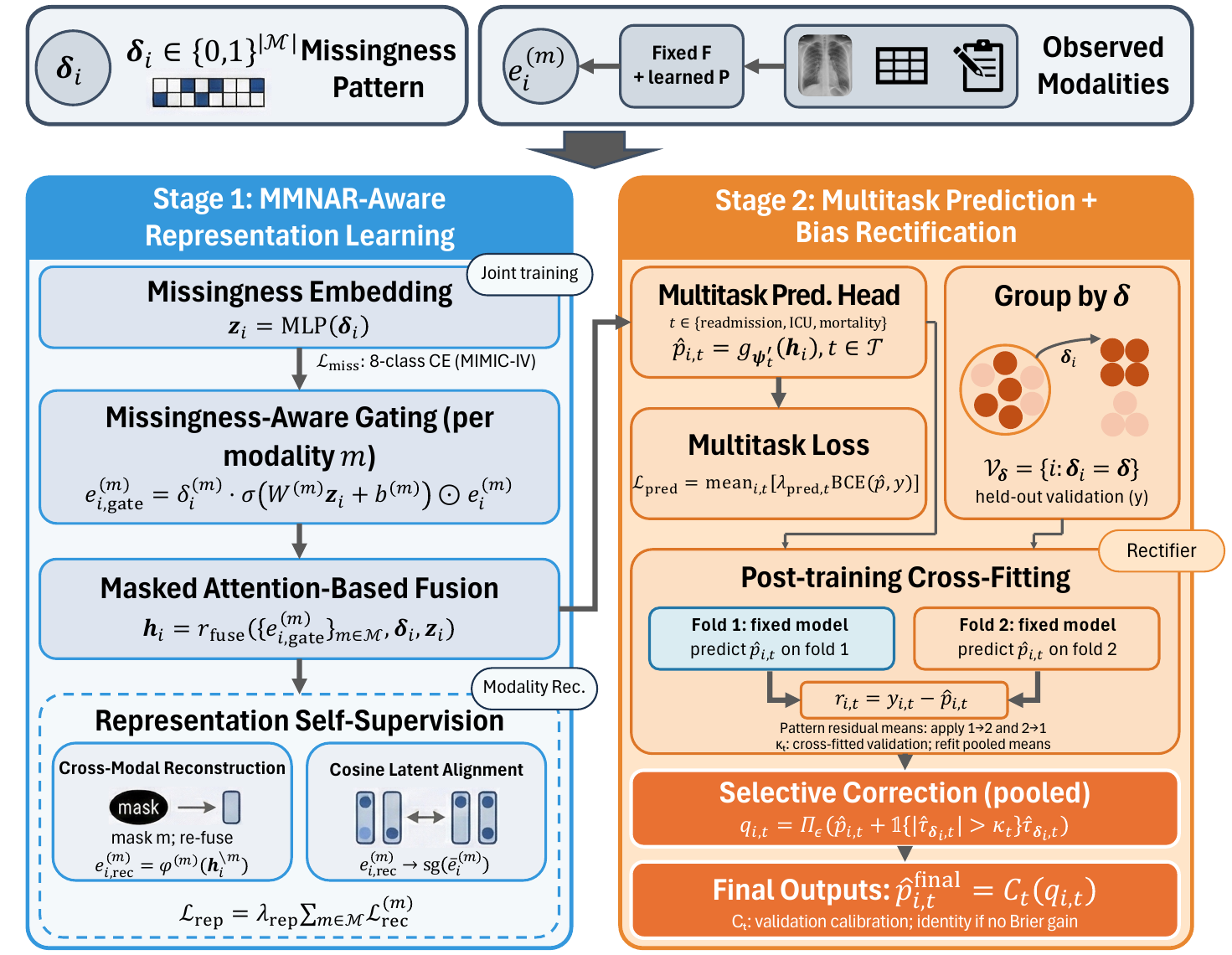}
    \caption{Overview of our \acronym{} method. Stage 1 learns patient representations through missingness-aware fusion and modality reconstruction with latent alignment. Stage 2 predicts clinical outcomes with multitask heads. Representation learning and base prediction are jointly optimized; cross-fitted rectification follows training.}
  \label{fig:flowchart}
\end{figure*}

\subsection{Preprocessing Multimodal Data}\label{method:preprocess}

For each observed modality $x_i^{(m)}$, we obtain an embedding $e_i^{(m)}=E^{(m)}(x_i^{(m)})\in\mathbb R^d$. In our MIMIC-IV implementation, $E^{(m)}$ comprises a separately trained supervised modality expert $F^{(m)}$ followed by a trainable projection $P^{(m)}$. The fusion network takes the expert's outcome logits as its modality-specific input representation:
\[
\bm\ell_i^{(m)}=F^{(m)}(x_i^{(m)}),\qquad
e_i^{(m)}=P^{(m)}(\tilde{\bm\ell}_i^{(m)}),
\]
where $\tilde{\bm\ell}_i^{(m)}$ denotes logits standardized using observed patients in the fusion fitting subset. The expert logits are held fixed during joint training, while the projections to the common $d$-dimensional space are learned. Missing logit inputs are set to zero, and the observation indicators remove their projected tokens through gating and attention masking.

The structured expert uses LightGBM on patient-level features; the imaging expert uses logistic regression on pretrained chest X-ray features \citep{sellergren2023generalized}; the discharge-summary expert fine-tunes BioClinical-ModernBERT \citep{sounack2025bioclinicalmodernbert}; and the radiology-report expert uses TF--IDF with logistic regression. The S, I, and R experts each supply three logits for the main outcomes. The T expert supplies four, including an auxiliary 90-day readmission logit. All four logit vectors are projected to $d=16$ before fusion. Expert training partitions and out-of-fold prediction are detailed in Appendix~\ref{appendix:protocol}. All inputs follow the same modality-eligibility rules used to define $\bm\delta_i$ (Appendix~\ref{appendix:data_preprocessing}).

\subsection{Patient Representation Learning}\label{method:rep-learning}

\subsubsection{MMNAR-Aware Modality Fusion}\label{method:mmnar}

The \textit{MMNAR-aware modality fusion} component integrates modality embeddings $\{e_i^{(m)}\}_{m \in \mathcal{M}_i}$ into a unified representation $\bm{h}_i$ (Equation \eqref{eqn:represenation-learning}). Unlike standard fusion strategies, this module explicitly treats missingness patterns $\bm{\delta}_i$ as structured contextual signals rather than random noise. The design preserves information in $\bm{\delta}_i$ alongside the modality embeddings $e_i^{(m)}$. The component proceeds in two main steps.

\textit{Step 1: Missingness-aware transformation}. We first compute a missingness embedding $\bm z_i=\mathrm{MLP}(\bm\delta_i)$, which transforms the binary observation pattern into a dense vector. A decoder maps $\bm z_i$ to a predicted distribution $\hat{\bm\pi}_i$ over pattern categories, with target $c_i=c(\bm\delta_i)$. For a training minibatch $\mathcal B$, the objective is
\[
\mathcal L_{\mathrm{miss}}=
\frac{\lambda_{\mathrm{miss}}}{|\mathcal B|}
\sum_{i\in\mathcal B}\mathrm{CE}(\hat{\bm\pi}_i,c_i),
\]
where $\mathrm{CE}$ denotes categorical cross-entropy and $\lambda_{\mathrm{miss}}$ controls the loss weight. In MIMIC-IV, S is always observed, so the decoder classifies the eight possible I/T/R configurations.

Each modality-specific embedding $e_i^{(m)}$ is then reweighted  according to the missingness embedding $\bm{z}_i$:
\[
e_{i,\text{gate}}^{(m)} = \delta_i^{(m)} \cdot \sigma\left(W^{(m)} \bm{z}_i + b^{(m)}\right) \odot e_i^{(m)},
\]
where $\odot$ denotes elementwise multiplication, $\sigma(\cdot)$ is a sigmoid gating function and $(W^{(m)}, b^{(m)})$ are modality-specific parameters. Missing modalities ($\delta_i^{(m)} = 0$) are imputed with zero, while observed ones are adaptively emphasized or attenuated based on $\bm{z}_i$.

\textit{Step 2: Attention-based fusion}. The gated embeddings $\{e_{i,\text{gate}}^{(m)}\}_{m \in \mathcal{M}}$ are then aggregated with a multi-head self-attention mechanism:
\[
\bm{h}_i = r_{\text{fuse}}\left(\{e_{i,\text{gate}}^{(m)}\}_{m \in \mathcal{M}},\bm\delta_i,\bm z_i\right).
\]
Missing modalities are masked in attention. The fusion map combines the observed modality representations with the missingness context, allowing each modality to be interpreted in relation to the others. The loss functions for learning $\bm{h}_i$ are described in the next subsection.

\subsubsection{Modality Reconstruction with Latent Alignment}
\label{method:recon}

The \textit{modality reconstruction with latent alignment} component encourages the fused patient representation $\bm h_i$ to retain information needed to recover masked modality representations and generalize across observation patterns.

\textit{Cross-modality reconstruction}. For each observed modality $m\in\mathcal M_i$, we randomly mask it when at least one other modality is available. Let $\mathcal A_m\subseteq\mathcal B$ contain the selected patients. Using the remaining modalities and the modified observation vector $\bm\delta_i^{\setminus m}$, we recompute the missingness embedding and partial representation $\bm h_i^{\setminus m}$ through the fusion process in Section~\ref{method:mmnar}. A modality-specific predictor reconstructs the excluded latent representation:
\[
e_{i,\mathrm{rec}}^{(m)}=\phi^{(m)}(\bm h_i^{\setminus m}).
\]

\textit{Latent alignment}. The target $\bar e_i^{(m)}$ is the reference embedding of the observed modality, held fixed when optimizing the reconstruction term. We align the normalized reconstruction with this target using
\[
\begin{split}
\mathcal L_{\mathrm{rec}}^{(m)}
={}&\frac{1}{|\mathcal A_m|}\sum_{i\in\mathcal A_m}
\Big[2-2\,\mathrm{sim}\big(
e_{i,\mathrm{rec}}^{(m)},\operatorname{sg}(\bar e_i^{(m)})
\big)\Big],
\end{split}
\]
where $\mathrm{sim}$ is cosine similarity and $\operatorname{sg}$ holds the target fixed in this loss. The loss is evaluated for modalities with at least two eligible masked observations and is zero otherwise. This objective encourages agreement between a reconstruction from the remaining modalities and the observed modality's reference representation.

Aggregating across modalities gives
\[
\mathcal L_{\mathrm{rep}}=
\lambda_{\mathrm{rep}}\sum_{m\in\mathcal M}\mathcal L_{\mathrm{rec}}^{(m)},
\]
where $\lambda_{\mathrm{rep}}$ controls the reconstruction objective. The reconstruction predictors are trained jointly with fusion and outcome prediction.

\subsection{Multitask Outcome Prediction with Rectifier}
\label{method:multitask_outcome}

The final part of \acronym{} is \textit{multitask outcome prediction with rectifier}. Treating the observation pattern $\bm\delta_i$ as the treatment assignment, we consider a simplified form of Equation~\eqref{eqn:multitask-outcome-prediction}, using an additive treatment-effect decomposition for each outcome $t$:
\begin{equation}\label{eqn:multitask-outcome-prediction-simplified}
    y_{i,t} =g_{\bm{\psi}^\prime_t}\big(\bm{h}_i\big) + \tau_{\bm{\delta}_i,t}+ \tilde\varepsilon_{i,t} \, . 
\end{equation}
The first term, $g_{\bm{\psi}^\prime_t}$, maps the shared representation $\bm h_i$ to a base outcome probability using a sigmoid output. Its parameters are shared across observation patterns, while the representation is shared across outcome tasks. In this additive model, the second term, $\tau_{\bm{\delta}_i,t}$, represents the treatment effect of observation pattern $\bm\delta_i$ on outcome $t$ that remains beyond the base prediction. This parameter captures the outcome contribution of modality assignment and associated care decisions not already represented by $g_{\bm\psi'_t}(\bm h_i)$. We operationalize it as the pattern-specific mean residual and estimate it on held-out validation data; $\tilde\varepsilon_{i,t}$ denotes the remaining residual. We estimate the predictor and treatment-effect component in two steps.

\textit{Step 1: Multitask prediction loss}.
We jointly train outcome-specific predictors using the shared representation $\bm h_i$. Writing the base probability as $\hat p_{i,t}=g_{\bm\psi'_t}(\bm h_i)$, the objective is
\[
\mathcal L_{\mathrm{pred}}=
\frac{1}{|\mathcal B|\,|\mathcal T|}
\sum_{i\in\mathcal B}\sum_{t\in\mathcal T}
\lambda_{\mathrm{pred},t}\,\mathrm{BCE}(\hat p_{i,t},y_{i,t}),
\]
where $\mathrm{BCE}$ denotes binary cross-entropy and $\lambda_{\mathrm{pred},t}$ is the task weight.

\textit{Step 2: Observation-pattern-specific rectifier}.
After training, we keep the encoder and task heads fixed and divide the held-out validation set into two disjoint folds, $\mathcal V_{\mathrm{val}}^{(1)}$ and $\mathcal V_{\mathrm{val}}^{(2)}$. For each pattern $\bm\delta$, let $\mathcal V_{\bm\delta}^{(k)}=\{i\in\mathcal V_{\mathrm{val}}^{(k)}:\bm\delta_i=\bm\delta\}$. For fold $k\in\{1,2\}$ and task $t$, we estimate the pattern-specific treatment-effect component as the average residual:
\[
\hat\tau^{(k)}_{\bm\delta,t}
=\frac{1}{|\mathcal V_{\bm\delta}^{(k)}|}
\sum_{i\in\mathcal V_{\bm\delta}^{(k)}}(y_{i,t}-\hat p_{i,t}).
\]
An absent pattern receives a correction of zero. The estimate from fold $k$ is applied to the complementary fold $\bar k$:
\[
\begin{split}
q^{\mathrm{cf}}_{i,t}=\Pi_\epsilon\!\Big(&\hat p_{i,t}
+\mathbf{1}\{|\hat\tau^{(k)}_{\bm\delta_i,t}|>\kappa_t\}
\hat\tau^{(k)}_{\bm\delta_i,t}\Big),\\[-2pt]
&i\in\mathcal V_{\mathrm{val}}^{(\bar k)},\quad\bar k\ne k,
\end{split}
\]
where $\Pi_\epsilon$ clips probabilities to $[\epsilon,1-\epsilon]$ for $0<\epsilon<1/2$, and $\kappa_t$ enables selective correction (Appendix~\ref{appendix:rectifier_example}). We select $\kappa_t$ using cross-fitted validation predictions, then estimate $\hat\tau_{\bm\delta,t}$ on the pooled validation residuals and apply the fixed corrections to test predictions.

The reported probability includes a validation-fitted calibration map $C_t$, taken to be the identity when calibration does not improve held-out Brier score:
\[
\begin{split}
\hat p^{\mathrm{final}}_{i,t}=C_t\!\left(\Pi_\epsilon\!\Big(
\hat p_{i,t}
+\mathbf{1}\{|\hat\tau_{\bm\delta_i,t}|>\kappa_t\}
\hat\tau_{\bm\delta_i,t}\Big)\right).
\end{split}
\]
Test labels are used only for evaluation.

Our rectifier is inspired by residual augmentation in semiparametric methods for missing data \citep{robins1994estimation,robins1995semiparametric} and prediction-powered inference \citep{angelopoulos2023prediction}. Here, the treatment is the modality observation pattern itself, which summarizes data-collection and care decisions. The augmentation term incorporates its treatment-effect component into the base prediction through the additive model above. Cross-fitting follows the sample-splitting principle used in semiparametric estimation \citep{chernozhukov2018double}: effects are estimated from residuals on one validation fold and evaluated on the other to limit overfitting.

\subsection{Joint Training Procedure}
\label{method:training}

We now describe the complete training procedure for the outcome model $y_{i,t} = f_{\bm{\theta}_t}(\bm{x}^{\text{obs}}_i, \bm\delta_i) + \varepsilon_{i,t}$. The total training objective combines three losses:
\[
\mathcal{L}_{\text{total}} = \underbrace{\mathcal{L}_{\text{miss}}}_{\text{Section}~\ref{method:mmnar}} + \underbrace{\mathcal{L}_{\text{rep}}}_{\text{Section}~\ref{method:recon}} + \underbrace{\mathcal{L}_{\text{pred}}}_{\text{Section}~\ref{method:multitask_outcome}} \, ,
\]
where $\mathcal{L}_{\text{miss}}$ learns missingness context, $\mathcal{L}_{\text{rep}}$ preserves cross-modal information, and $\mathcal{L}_{\text{pred}}$ optimizes outcome prediction.

The self-supervised terms, $\mathcal{L}_{\text{miss}}+\mathcal{L}_{\text{rep}}$, update the representation encoder $\bm\eta$ in Equation~\eqref{eqn:represenation-learning}. The supervised prediction loss $\mathcal{L}_{\text{pred}}$ updates both $\bm\eta$ and the task-specific heads $\bm\psi_t^\prime$ in Equation~\eqref{eqn:multitask-outcome-prediction-simplified}. These objectives jointly train the common-space projections, fusion network, reconstruction predictors, and task-specific heads, with the supervised expert logits held fixed in MIMIC-IV. Stage 1 and Stage 2 denote functional components of the model.

After training, we estimate the observation-pattern-specific treatment-effect components $\{\hat\tau_{\bm\delta,t}\}$ using cross-fitted validation predictions and pooled residuals. The rectifier incorporates these effects, beyond those captured by the shared representation, into the final outcome probabilities.

Putting everything together, $\bm\psi_t=(\bm\psi_t^\prime,\{\hat\tau_{\bm\delta,t}\},\kappa_t,C_t)$ collects the task head and post-training adjustments, giving $\bm\theta_t=(\bm\eta,\bm\psi_t)$ as in Section~\ref{methods}. Loss weights and the evaluation protocol are provided in Appendix~\ref{appendix:protocol}.

\section{Experiments}
\label{experiment}

\subsection{Datasets and Preprocessing}
\label{experiment:datasets}

To evaluate \acronym{} across different healthcare settings, we conduct experiments on two large-scale clinical datasets. 

\paragraph{MIMIC-IV Dataset}

We link structured EHRs from MIMIC-IV v3.1 \cite{johnson2024mimic} with clinical notes from MIMIC-IV-Note v2.2 and chest radiographs from MIMIC-CXR. The three MIMIC-IV tasks share a cohort of 20{,}000 adults ($\geq$18 years) discharged alive from their first recorded hospital admission, with one index hospitalization per patient. ICU admission is not required for cohort eligibility. Subsequent admissions provide follow-up information for readmission and ICU outcomes.

Among these patients, 12,970 (64.85\%) have discharge summaries, 12,793 (63.97\%) have radiology reports, and all have structured data. Overall, 15,261 (76.31\%) have at least one text modality, 1,367 (6.84\%) have eligible CXRs, and 1,293 (6.47\%) have all four modalities. The cohort includes 2,749 readmission events (13.7\%), 689 subsequent ICU admissions (3.4\%), and 639 deaths within 90 days after discharge (3.2\%). Counts and outcome occurance rates for the eight possible modality patterns are reported in Table~\ref{tab:pattern_outcomes}; seven patterns are observed in this cohort. For downstream modeling, we encode each patient's modality availability as a binary vector.

All three tasks use the common discharge-based inputs described in Appendix~\ref{appendix:data_preprocessing}. Discharge-summary embeddings and availability indicators are included whenever eligible notes are present, including for mortality prediction.

\paragraph{eICU Dataset}

We further evaluate on the eICU Collaborative Research Database \cite{pollard_eicu_2018}, which spans 208 hospitals across the United States. This multi-center dataset provides an additional evaluation setting with heterogeneous institutions and care practices.

Since eICU consists primarily of structured data, we define \emph{virtual modalities}. Features are grouped into 10 clinical categories: Demographics, Admission, APACHE, Diagnosis, Medication, Laboratory Values, Vital Signs, Respiratory, Fluid Balance, and Comorbidities. For each virtual modality $m$, which consists of a group of related features (e.g., all laboratory measurements or all vitals), we treat it as missing for patient $i$ if more than 80\% of the features within that group are NaN. This virtual modality approach follows established practices for handling heterogeneous EHR data, where clinical features are systematically grouped into meaningful categories to enable effective multimodal learning \cite{wang_multimodal_2024,li_integrating_2022}.

Cohort selection, preprocessing, and definitions of the eICU virtual modalities are provided in Appendix~\ref{appendix:data_preprocessing}.

\paragraph{Clinical Tasks} We evaluate three clinical outcomes in MIMIC-IV, and readmission and mortality in eICU: (1) \textbf{30-day Hospital Readmission}: Predicting readmission within 30 days post-discharge; (2) \textbf{Post-discharge ICU Admission}: Predicting ICU admission within 90 days post-discharge; (3) \textbf{Mortality}: Predicting 90-day post-discharge all-cause mortality in MIMIC-IV and in-hospital mortality in eICU. All three MIMIC-IV outcomes are defined relative to index hospital discharge among discharge survivors and use a common retrospective observation window. MIMIC-IV mortality labels use recorded patient death dates; calendar-day boundaries and labeling details are provided in Appendix~\ref{appendix:data_preprocessing}.

\subsection{Baselines and Evaluation Protocol}
\label{experiment:baselines}

We compare \acronym{} with 13 representative methods, spanning three major paradigms in multimodal learning. The second paradigm is explicitly designed to handle missing data.

\begin{table*}[ht]
\centering
\small
\setlength{\tabcolsep}{1.5pt}
\caption{Performance comparison across datasets and clinical tasks. Point estimates with 95\% confidence intervals in subscript brackets; best estimates in \textbf{bold}. AUPRC denotes average precision. Higher AUC/AUPRC and lower Brier are better.}
\label{tab:main_results}
\resizebox{\textwidth}{!}{%
\begin{tabular}{l|ccc|ccc|ccc}
\toprule
\multirow{3}{*}{\textbf{Model}} & \multicolumn{9}{c}{\textbf{MIMIC-IV Dataset}} \\
\cmidrule{2-10}
& \multicolumn{3}{c|}{\textbf{30-day Readmission}} & \multicolumn{3}{c|}{\textbf{Post-discharge ICU}} & \multicolumn{3}{c}{\textbf{90-day Mortality}} \\
& \textbf{AUC} & \textbf{AUPRC} & \textbf{Brier} & \textbf{AUC} & \textbf{AUPRC} & \textbf{Brier} & \textbf{AUC} & \textbf{AUPRC} & \textbf{Brier} \\
\midrule
CM-AE & \resultci{0.675}{0.651}{0.699} & \resultci{0.275}{0.243}{0.311} & \resultci{0.1123}{0.1058}{0.1190} & \resultci{0.777}{0.740}{0.812} & \resultci{0.112}{0.084}{0.157} & \resultci{0.0321}{0.0273}{0.0371} & \resultci{0.885}{0.857}{0.913} & \resultci{0.424}{0.339}{0.514} & \resultci{0.0249}{0.0210}{0.0291} \\
MT & \resultci{0.698}{0.675}{0.721} & \resultci{0.282}{0.249}{0.321} & \resultci{0.1104}{0.1035}{0.1174} & \resultci{0.803}{0.766}{0.837} & \resultci{0.144}{0.106}{0.203} & \resultci{0.0315}{0.0267}{0.0365} & \resultci{0.905}{0.879}{0.927} & \resultci{0.429}{0.338}{0.519} & \resultci{0.0247}{0.0209}{0.0289} \\
SMIL & \resultci{0.693}{0.669}{0.717} & \resultci{0.286}{0.252}{0.324} & \resultci{0.1107}{0.1041}{0.1176} & \resultci{0.806}{0.773}{0.836} & \resultci{0.132}{0.097}{0.183} & \resultci{0.0318}{0.0269}{0.0368} & \resultci{0.920}{0.898}{0.939} & \resultci{0.434}{0.348}{0.529} & \resultci{0.0248}{0.0208}{0.0293} \\
GRAPE & \resultci{0.718}{0.694}{0.740} & \resultci{0.306}{0.270}{0.345} & \resultci{0.1085}{0.1016}{0.1155} & \resultci{0.828}{0.797}{0.856} & \resultci{0.142}{0.110}{0.195} & \resultci{0.0317}{0.0272}{0.0364} & \resultci{0.939}{0.923}{0.955} & \resultci{0.466}{0.382}{0.556} & \resultci{0.0238}{0.0202}{0.0276} \\
HGMF & \resultci{0.704}{0.678}{0.728} & \resultci{0.341}{0.302}{0.384} & \resultci{0.1072}{0.1005}{0.1142} & \resultci{0.792}{0.756}{0.826} & \resultci{0.161}{0.112}{0.232} & \resultci{0.0313}{0.0267}{0.0362} & \resultci{0.928}{0.910}{0.946} & \resultci{0.476}{0.392}{0.560} & \resultci{0.0236}{0.0199}{0.0274} \\
M3Care & \resultci{0.697}{0.671}{0.723} & \resultci{0.371}{0.329}{0.411} & \resultci{0.1058}{0.0990}{0.1128} & \resultci{0.800}{0.764}{0.833} & \resultci{0.181}{0.128}{0.244} & \resultci{0.0311}{0.0265}{0.0358} & \resultci{0.933}{0.911}{0.952} & \resultci{0.493}{0.406}{0.582} & \resultci{0.0229}{0.0193}{0.0266} \\
COM & \resultci{0.741}{0.718}{0.763} & \resultci{0.360}{0.320}{0.401} & \resultci{0.1046}{0.0982}{0.1113} & \resultci{0.809}{0.774}{0.846} & \resultci{0.187}{0.134}{0.254} & \resultci{0.0307}{0.0259}{0.0357} & \resultci{0.950}{0.935}{0.964} & \resultci{0.516}{0.432}{0.605} & \resultci{0.0223}{0.0188}{0.0260} \\
DrFuse & \resultci{0.735}{0.711}{0.757} & \resultci{0.348}{0.310}{0.390} & \resultci{0.1058}{0.0988}{0.1128} & \resultci{0.845}{0.813}{0.875} & \resultci{0.228}{0.170}{0.305} & \resultci{0.0300}{0.0252}{0.0346} & \resultci{0.944}{0.929}{0.957} & \resultci{0.529}{0.445}{0.612} & \resultci{0.0222}{0.0189}{0.0259} \\
MissModal & \resultci{0.701}{0.674}{0.725} & \resultci{0.346}{0.308}{0.387} & \resultci{0.1074}{0.1008}{0.1142} & \resultci{0.819}{0.786}{0.852} & \resultci{0.188}{0.138}{0.256} & \resultci{0.0308}{0.0262}{0.0354} & \resultci{0.931}{0.908}{0.951} & \resultci{0.496}{0.408}{0.584} & \resultci{0.0230}{0.0193}{0.0269} \\
FLEXGEN-EHR & \resultci{0.735}{0.713}{0.756} & \resultci{0.347}{0.308}{0.389} & \resultci{0.1058}{0.0989}{0.1130} & \resultci{0.830}{0.793}{0.862} & \resultci{0.212}{0.158}{0.284} & \resultci{0.0302}{0.0257}{0.0351} & \resultci{0.941}{0.926}{0.955} & \resultci{0.492}{0.409}{0.581} & \resultci{0.0236}{0.0201}{0.0273} \\
MUSE+ & \resultci{0.702}{0.678}{0.727} & \resultci{0.318}{0.280}{0.359} & \resultci{0.1083}{0.1011}{0.1156} & \resultci{0.813}{0.774}{0.848} & \resultci{0.203}{0.149}{0.271} & \resultci{0.0304}{0.0257}{0.0352} & \resultci{0.932}{0.915}{0.948} & \resultci{0.447}{0.364}{0.535} & \resultci{0.0244}{0.0207}{0.0285} \\
GRU-D & \resultci{0.648}{0.623}{0.674} & \resultci{0.242}{0.213}{0.276} & \resultci{0.1140}{0.1067}{0.1214} & \resultci{0.773}{0.732}{0.810} & \resultci{0.108}{0.082}{0.148} & \resultci{0.0322}{0.0273}{0.0373} & \resultci{0.814}{0.777}{0.850} & \resultci{0.251}{0.185}{0.327} & \resultci{0.0292}{0.0249}{0.0336} \\
Raindrop & \resultci{0.731}{0.707}{0.752} & \resultci{0.336}{0.298}{0.379} & \resultci{0.1064}{0.0993}{0.1135} & \resultci{0.833}{0.797}{0.864} & \resultci{0.173}{0.133}{0.233} & \resultci{0.0309}{0.0263}{0.0359} & \resultci{0.936}{0.918}{0.953} & \resultci{0.463}{0.380}{0.559} & \resultci{0.0244}{0.0208}{0.0281} \\
\midrule
\textbf{\acronym{}} & \resultci{\textbf{0.773}}{0.751}{0.794} & \resultci{\textbf{0.418}}{0.374}{0.459} & \resultci{\textbf{0.0998}}{0.0931}{0.1066} & \resultci{\textbf{0.868}}{0.838}{0.896} & \resultci{\textbf{0.320}}{0.242}{0.405} & \resultci{\textbf{0.0278}}{0.0235}{0.0324} & \resultci{\textbf{0.957}}{0.943}{0.969} & \resultci{\textbf{0.570}}{0.484}{0.661} & \resultci{\textbf{0.0207}}{0.0175}{0.0242} \\
\bottomrule
\end{tabular}
}

\vspace{0.3cm}

\resizebox{0.8\textwidth}{!}{%
\begin{tabular}{l|ccc|ccc}
\toprule
\multirow{3}{*}{\textbf{Model}} & \multicolumn{6}{c}{\textbf{eICU Dataset}} \\
\cmidrule{2-7}
& \multicolumn{3}{c|}{\textbf{30-day Readmission}} & \multicolumn{3}{c}{\textbf{In-hospital Mortality}} \\
& \textbf{AUC} & \textbf{AUPRC} & \textbf{Brier} & \textbf{AUC} & \textbf{AUPRC} & \textbf{Brier} \\
\midrule
CM-AE & \resultci{0.734}{0.695}{0.773} & \resultci{0.138}{0.101}{0.190} & \resultci{0.0372}{0.0322}{0.0427} & \resultci{0.859}{0.837}{0.880} & \resultci{0.485}{0.427}{0.541} & \resultci{0.0568}{0.0514}{0.0625} \\
MT & \resultci{0.749}{0.709}{0.788} & \resultci{0.166}{0.122}{0.229} & \resultci{0.0365}{0.0315}{0.0421} & \resultci{0.884}{0.865}{0.902} & \resultci{0.523}{0.471}{0.577} & \resultci{0.0544}{0.0494}{0.0600} \\
SMIL & \resultci{0.745}{0.709}{0.781} & \resultci{0.136}{0.099}{0.193} & \resultci{0.0372}{0.0320}{0.0429} & \resultci{0.880}{0.860}{0.899} & \resultci{0.578}{0.528}{0.629} & \resultci{0.0510}{0.0458}{0.0565} \\
GRAPE & \resultci{0.770}{0.734}{0.804} & \resultci{0.172}{0.125}{0.234} & \resultci{0.0364}{0.0315}{0.0419} & \resultci{0.898}{0.878}{0.915} & \resultci{0.587}{0.536}{0.639} & \resultci{0.0502}{0.0453}{0.0554} \\
HGMF & \resultci{0.755}{0.718}{0.789} & \resultci{0.160}{0.116}{0.216} & \resultci{0.0368}{0.0319}{0.0423} & \resultci{0.881}{0.864}{0.899} & \resultci{0.498}{0.446}{0.553} & \resultci{0.0556}{0.0502}{0.0609} \\
M3Care & \resultci{0.760}{0.727}{0.797} & \resultci{0.133}{0.101}{0.189} & \resultci{0.0373}{0.0320}{0.0431} & \resultci{0.911}{0.893}{0.927} & \resultci{0.610}{0.557}{0.666} & \resultci{0.0478}{0.0432}{0.0526} \\
COM & \resultci{0.771}{0.736}{0.809} & \resultci{0.182}{0.137}{0.247} & \resultci{0.0362}{0.0311}{0.0416} & \resultci{0.904}{0.887}{0.920} & \resultci{0.605}{0.555}{0.654} & \resultci{0.0490}{0.0440}{0.0543} \\
DrFuse & \resultci{0.780}{0.746}{0.814} & \resultci{0.180}{0.136}{0.247} & \resultci{0.0362}{0.0314}{0.0414} & \resultci{0.918}{0.902}{0.932} & \resultci{0.636}{0.583}{0.684} & \resultci{0.0467}{0.0416}{0.0515} \\
MissModal & \resultci{0.746}{0.707}{0.783} & \resultci{0.144}{0.108}{0.201} & \resultci{0.0369}{0.0318}{0.0424} & \resultci{0.885}{0.866}{0.904} & \resultci{0.548}{0.492}{0.599} & \resultci{0.0528}{0.0479}{0.0578} \\
FLEXGEN-EHR & \resultci{0.743}{0.705}{0.782} & \resultci{0.144}{0.107}{0.193} & \resultci{0.0371}{0.0321}{0.0426} & \resultci{0.896}{0.880}{0.913} & \resultci{0.585}{0.533}{0.634} & \resultci{0.0506}{0.0454}{0.0561} \\
MUSE+ & \resultci{0.801}{0.766}{0.834} & \resultci{0.199}{0.149}{0.260} & \resultci{0.0358}{0.0307}{0.0413} & \resultci{0.928}{0.913}{0.942} & \resultci{0.673}{0.626}{0.720} & \resultci{0.0438}{0.0392}{0.0486} \\
GRU-D & \resultci{0.700}{0.653}{0.742} & \resultci{0.154}{0.111}{0.219} & \resultci{0.0370}{0.0318}{0.0427} & \resultci{0.840}{0.817}{0.861} & \resultci{0.423}{0.371}{0.479} & \resultci{0.0607}{0.0550}{0.0660} \\
Raindrop & \resultci{0.737}{0.697}{0.778} & \resultci{0.187}{0.139}{0.258} & \resultci{0.0363}{0.0315}{0.0416} & \resultci{0.868}{0.846}{0.888} & \resultci{0.479}{0.424}{0.537} & \resultci{0.0571}{0.0516}{0.0628} \\
\midrule
\textbf{\acronym{}} & \resultci{\textbf{0.819}}{0.784}{0.851} & \resultci{\textbf{0.238}}{0.181}{0.307} & \resultci{\textbf{0.0348}}{0.0300}{0.0400} & \resultci{\textbf{0.946}}{0.933}{0.957} & \resultci{\textbf{0.765}}{0.721}{0.805} & \resultci{\textbf{0.0359}}{0.0321}{0.0404} \\
\bottomrule
\end{tabular}
}
\label{tab:eicu_results}
\end{table*}

The baselines cover three groups: traditional multimodal methods, including CM-AE~\citep{cmae}, MT~\citep{ma2022mt}, GRAPE~\citep{you2020grape}, and HGMF~\citep{chen2020hgmf}; methods designed for missing modalities, including SMIL~\citep{ma2021smil}, M3Care~\citep{zhang2022m3care}, COM~\citep{qian_com_2023}, DrFuse~\citep{yao_drfuse_2024}, MissModal~\citep{lin2023missmodal}, FLEXGEN-EHR~\citep{he_flexible_2024}, and MUSE+~\citep{wu_multimodal_2024}; and irregular time-series methods GRU-D~\citep{che2018gru-d} and Raindrop~\citep{zhang2022graph}.

\begin{table*}[ht]
\centering
\renewcommand{\arraystretch}{1.1}
\caption{Component ablation on MIMIC-IV and eICU. MMNAR denotes missingness-aware fusion; MR denotes modality reconstruction with latent alignment; Rectifier includes pattern-specific correction and final probability calibration. $\Delta$AUC is the increment over the preceding row, computed before rounding.}
\label{tab:ablation_study}
\resizebox{\textwidth}{!}{
\LARGE
\begin{tabular}{l|ccc|ccc|ccc|ccc|ccc}
\toprule
\multirow{3}{*}{\textbf{Component}}
& \multicolumn{9}{c|}{\textbf{MIMIC-IV Dataset}}
& \multicolumn{6}{c}{\textbf{eICU Dataset}} \\
\cmidrule(lr){2-10} \cmidrule(lr){11-16}
& \multicolumn{3}{c|}{\textbf{30-day Readmission}}
& \multicolumn{3}{c|}{\textbf{Post-discharge ICU}}
& \multicolumn{3}{c|}{\textbf{90-day Mortality}}
& \multicolumn{3}{c|}{\textbf{30-day Readmission}}
& \multicolumn{3}{c}{\textbf{In-hospital Mortality}} \\
\cmidrule(lr){2-4} \cmidrule(lr){5-7} \cmidrule(lr){8-10} \cmidrule(lr){11-13} \cmidrule(lr){14-16}
& \textbf{AUC} & \textbf{AUPRC} & $\boldsymbol\Delta$\textbf{AUC}
& \textbf{AUC} & \textbf{AUPRC} & $\boldsymbol\Delta$\textbf{AUC}
& \textbf{AUC} & \textbf{AUPRC} & $\boldsymbol\Delta$\textbf{AUC}
& \textbf{AUC} & \textbf{AUPRC} & $\boldsymbol\Delta$\textbf{AUC}
& \textbf{AUC} & \textbf{AUPRC} & $\boldsymbol\Delta$\textbf{AUC} \\
\midrule
Basic Baseline
& 0.706 & 0.310 & \textemdash
& 0.788 & 0.138 & \textemdash
& 0.870 & 0.307 & \textemdash
& 0.759 & 0.171 & \textemdash
& 0.901 & 0.585 & \textemdash \\
+ MMNAR
& 0.735 & 0.369 & 0.029
& 0.823 & 0.250 & 0.036
& 0.904 & 0.472 & 0.034
& 0.787 & 0.205 & 0.028
& 0.923 & 0.675 & 0.022 \\
+ MR
& 0.758 & 0.393 & 0.023
& 0.849 & 0.305 & 0.025
& 0.933 & 0.513 & 0.029
& 0.804 & 0.224 & 0.017
& 0.935 & 0.723 & 0.012 \\
+ Rectifier
& \textbf{0.773} & \textbf{0.418} & 0.015
& \textbf{0.868} & \textbf{0.320} & 0.019
& \textbf{0.957} & \textbf{0.570} & 0.024
& \textbf{0.819} & \textbf{0.238} & 0.015
& \textbf{0.946} & \textbf{0.765} & 0.011 \\
\bottomrule
\end{tabular}
}
\end{table*}

\paragraph{Evaluation Protocol} For MIMIC-IV, all three outcomes use the same discharge-survivor cohort, input eligibility rules, and patient split: 13,600 development patients, 2,400 validation patients, and 4,000 test patients. An additional 170,588 eligible patients outside the fixed cohort support training of modality representations. Validation and test labels are excluded from representation training, and validation data are used for rectification and calibration.

All MIMIC-IV methods share the eligible patient population, outcome definitions, retrospective input windows, and training population. Baselines use a common internal early-stopping set and at most four predeclared configurations per method. Configurations are selected on the 1,200-patient meta-selection partition by AP, with AUC breaking ties; Platt calibration is fitted only on the separate 1,200-patient meta-calibration partition. Each method is frozen before test scoring. Appendix~\ref{appendix:protocol} details the baseline protocol and the model-specific training procedure for \acronym{}.

Results are reported as point estimates with 95\% confidence intervals. For MIMIC-IV, \acronym{} uses a three-seed fusion ensemble: task logits are averaged before the sigmoid transformation, followed by validation-fitted rectification and calibration. Metrics are computed from the ensemble's final test probabilities. All MIMIC-IV methods use the same 2,000 patient-level bootstrap resamples (seed 20261022), enabling paired comparisons on identical resampled patients. Predictions remain fixed, so the intervals quantify test-patient sampling uncertainty conditional on the fitted models. Additional details on the cohort and evaluation protocol are provided in Appendices~\ref{appendix:data_preprocessing} and~\ref{appendix:protocol}.

\subsection{Main Results}
\label{experiment:main}

Table~\ref{tab:main_results} reports results across both datasets and all clinical tasks, using Area Under the ROC Curve (AUC), precision--recall performance, and Brier score as evaluation metrics. Precision--recall performance is summarized using average precision (AP), labeled AUPRC in the tables.

\paragraph{Performance Analysis} Performance is improved by \acronym{} across all tasks and metrics.
On MIMIC-IV, readmission AUC rises from $0.741$ with COM to $0.773$ ($\mathbf{+4.3\%}$), while ICU admission AUC rises from $0.845$ with DrFuse to $0.868$ ($\mathbf{+2.7\%}$).
For 90-day post-discharge mortality, AUC improves from $0.950$ with COM to $0.957$ ($\mathbf{+0.7\%}$).
On eICU, readmission AUC increases from $0.801$ with MUSE+ to $0.819$ ($\mathbf{+2.2\%}$), while mortality AUC rises from $0.928$ with MUSE+ to $0.946$ ($\mathbf{+1.9\%}$). Percentages denote relative AUC gains over the strongest AUC baseline.

These gains are accompanied by higher AUPRC and lower Brier scores, indicating improvements in precision--recall performance and lower squared error in predicted outcome probabilities. The results across two healthcare datasets support the applicability of our MMNAR-aware framework.

\paragraph{Robustness and Validation Studies} We conduct supplementary analyses to evaluate robustness and stability.

First, performance across modality configurations (Table~\ref{tab:modality_configs}, Appendix~\ref{appendix:performance_modality}) shows discrimination in the larger incomplete-modality groups, with the highest AUC and AUPRC in the complete-modality group. Estimates for rare patterns are unstable or undefined when only one outcome class is present. Overall comparisons with imputation methods favor our framework across both tasks and all three metrics (Table~\ref{tab:imputation_comparison}).

Second, hyperparameter sensitivity analysis for MIMIC-IV readmission (Table~\ref{tab:hyperparam_sensitivity}, Appendix~\ref{appendix:hyperparameter}) shows a maximum reported AUC decrease of $0.007$, with relative sensitivity of $0.86\%$ across the tested settings.

Finally, embedding analysis (Appendix~\ref{appendix:embedding}) shows that learned missingness-embedding norms vary across observed patterns. Their relationship with readmission rates is descriptive and sensitive to the very small imaging subgroups.

\subsection{Component Ablation Studies}
\label{experiment:ablation_study}

Table~\ref{tab:ablation_study} presents ablation results on MIMIC-IV and eICU demonstrating each component's contribution. Starting from a multimodal baseline with standard feature concatenation, we progressively add: (1) MMNAR-aware fusion with $\mathcal L_{\mathrm{miss}}$; (2) modality reconstruction with $\mathcal L_{\mathrm{rep}}$; and (3) the rectifier mechanism, including final probability calibration.

MMNAR-aware fusion provides the largest incremental AUC gain for every task on both datasets, indicating that observation patterns contain useful signals beyond standard fusion. Modality reconstruction delivers further gains, supporting the value of cross-modal latent alignment. The rectifier provides additional improvements: for ICU admission, AUC increases from $0.849$ to $0.868$, while AUPRC rises from $0.305$ to $0.320$.

On eICU, the full framework increases AUC from $0.759$ to $0.819$ for readmission and from $0.901$ to $0.946$ for in-hospital mortality. The rectifier adds $0.015$ and $0.011$ AUC over the MR variant, respectively, while AUPRC rises from $0.224$ to $0.238$ and from $0.723$ to $0.765$.

The complete framework achieves higher AUC and AUPRC than the basic baseline across all tasks on both datasets.

\section{Related Work}
\label{related_work}

Multimodal methods address incomplete inputs through reconstruction and missing-modality training \cite{cmae,ma2021smil}. Healthcare fusion methods, including MUSE+ and DrFuse, adapt multimodal learning to sparse clinical records \cite{wu_multimodal_2024,yao_drfuse_2024}. Temporal models also exploit missingness in longitudinal measurements \cite{che2018gru-d,zhang2022graph}. Our framework conditions representation learning and outcome prediction on modality observation patterns, treating their availability as informative context.

Graph-based fusion represents relationships among heterogeneous modalities \cite{you2020grape,chen2020hgmf}. Contrastive approaches provide complementary ways to learn from partial records \cite{qian_com_2023,lin2023missmodal}. These approaches motivate reconstruction and fusion for incomplete inputs. Our emphasis is the clinical information in modality assignment, incorporated into representation learning and outcome correction.

Our approach also connects to latent-factor methods for endogenous missingness \cite{xiong2023,duan2024factor,duan2024target} and semiparametric residual correction \cite{robins1994estimation,robins1995semiparametric}. Multitask learning shares information across related outcomes \cite{bengio2013representation}, while negative transfer remains a concern as the task set grows \cite{wu2020understanding,yang2025precise}. We discuss these connections and further related work in Appendix~\ref{appendix:relatedwork}.

\section{Conclusion}
We introduce \acronym{}, a causal multimodal framework that explicitly models MMNAR in clinical data. The framework combines missingness-aware fusion, cross-modal reconstruction, and multitask prediction with rectification to learn robust patient representations. Evaluations on MIMIC-IV and eICU show consistent improvements over $13$ baselines, with relative AUC gains on MIMIC-IV of 4.3\% for readmission and 2.7\% for ICU admission. This work highlights the importance of treating missingness as structured signal and offers a principled approach for patient representation learning under realistic data constraints.

\clearpage
\section{Limitations}
\label{limitations}

Our framework, though effective under Modality Missing-Not-at-Random (MMNAR) conditions, has several limitations. Its generalizability beyond MIMIC-IV and eICU remains uncertain without broader international validation, and reliance on proxy assumptions may not fully capture underlying causal mechanisms. Performance under random missingness or missingness driven by non-clinical factors remains to be assessed, and rare modality patterns limit subgroup-level conclusions. The multi-component architecture, while modular, introduces complexity, computational demands, and interpretability challenges that may hinder deployment in low-resource settings. Future work should focus on lightweight variants, improved interpretability, and wider cross-institutional evaluation. Potential risks about the methodology can be found in Appendix \ref{appendix:potential}.

\bibliography{custom}

\clearpage
\appendix
\onecolumn

\section{Rectifier Mechanism: Numerical Example}
\label{appendix:rectifier_example}

To illustrate the rectifier mechanism in Section~\ref{method:multitask_outcome}, consider a hypothetical pattern-specific correction for 30-day readmission.

\paragraph{Setup}
\begin{itemize}
  \item Observation pattern $\bm\delta$: structured data only (S)
  \item Task $t$: 30-day readmission
  \item Held-out residual estimate: $\hat\tau_{\bm\delta,t}=+0.08$
  \item Selected threshold: $\kappa_t=0.05$
  \item Base probability for a new patient: $\hat p_{i,t}=0.30$
\end{itemize}

\paragraph{Rectification}
Since $|\hat\tau_{\bm\delta,t}|>\kappa_t$, the additive correction gives
\[
q_{i,t}=\Pi_\epsilon(0.30+0.08)=0.38.
\]
With identity calibration, $\hat p^{\mathrm{final}}_{i,t}=0.38$; otherwise the validation-fitted map $C_t$ is applied to $q_{i,t}$.

\paragraph{Interpretation}
In this hypothetical example, the estimated treatment-effect component for the S observation pattern is $+8$ percentage points beyond the base prediction. Incorporating this effect through the additive rectifier increases predicted risk from 30\% to 38\%, correcting underprediction in the S subgroup.

\section{Additional Related Work}
\label{appendix:relatedwork}
We expand on the multimodal, causal, and multitask learning literature summarized in Section~\ref{related_work}, and review related work on medical foundation models, interpretability, trial design, and multimodal data synthesis.

\paragraph{Multimodal Learning with Missing Modalities.}
Our work is most closely related to the growing literature on multimodal representation learning, and in particular to methods that address missing modalities. Early approaches, such as CM-AE~\cite{cmae}, introduced autoencoder-based frameworks for cross-modal imputation. Subsequent methods developed more sophisticated modality-specific encoders with fusion mechanisms, including multimodal Transformers~\cite{ma2022mt}, graph-based aggregation (GRAPE~\cite{you2020grape}), and heterogeneous graph-based fusion (HGMF~\cite{chen2020hgmf}). These approaches address learning from incomplete inputs. A complementary line of work makes robustness to missing modalities a primary modeling objective. Examples include SMIL~\cite{ma2021smil}, COM~\cite{qian_com_2023}, and MissModal~\cite{lin2023missmodal}. These approaches directly relate to our setting by treating missingness as a key modeling consideration.

Within this stream, several methods have been developed specifically for healthcare applications, including M3Care~\cite{zhang2022m3care}, MUSE+~\cite{wu_multimodal_2024}, FLEXGEN-EHR~\cite{he_flexible_2024}, and DrFuse~\cite{yao_drfuse_2024}, which achieve strong performance under real-world modality sparsity. In parallel, temporal models such as GRU-D~\cite{che2018gru-d} and Raindrop~\cite{zhang2022graph} are developed to capture implicit temporal missingness patterns. These approaches address prediction from incomplete records. Our \acronym{} focuses on conditioning representation learning and residual outcome correction on the observed modality pattern.

\paragraph{Causal Learning and MNAR Mechanisms.}
Our work also connects to the growing literature at the intersection of causal inference and machine learning. We share an interest in using causal perspectives to improve prediction in observational settings \cite{kuang_stable_2018,kuang2020stable}. More closely, our work relates to research on uncovering latent structures when data are missing not at random, where missingness is endogenously driven by unobserved factors \cite{xiong2023,duan2024factor,duan2024target}. Finally, our rectifier mechanism parallels semiparametric approaches for handling missing data \cite{robins1994estimation,robins1995semiparametric}, as it uses held-out residuals to correct prediction bias associated with observation patterns.

\paragraph{Multitask Learning and Negative Transfer.}
Finally, our work also relates to the literature on multitask learning, which leverages commonalities across related prediction tasks to share statistical strength \cite{bengio2013representation}. In our setting, outcome tasks share a common representation, but as the number of outcomes grows, negative transfer may occur, where shared representations harm accuracy for certain tasks \cite{wu2020understanding,yang2025precise}, making its mitigation an important future direction. Recent advances in task modeling \cite{li2023boosting,li2023identification}, adaptive and scalable fine-tuning for individual tasks \cite{li2024scalable,li2024scalable2,li2025efficient} offer promising approaches to mitigate negative transfer by dynamically controlling when and how knowledge is shared across tasks.

\paragraph{Medical Foundation Models and Prompting.}
Recent innovations in medical foundation models extend multimodal learning to large-scale LLMs. 
Med-MLLM~\cite{liu_medical_2023} exemplifies a multimodal LLM that generalizes across visual and textual clinical data, achieving robust performance in few-shot pandemic prediction tasks. 
CHiLL~\cite{mcinerney_chill_2023} leverages zero-shot prompting of LLMs to generate interpretable features from free-text notes, showing that clinically meaningful representations can be constructed without manual engineering. 
These efforts highlight complementary strategies for scalability beyond task-specific architectures.

\paragraph{Interpretability and Transparency.}
Model transparency remains essential for clinician adoption. 
Stenwig et al.~\cite{stenwig_comparative_2022} and Nohara et al.~\cite{nohara_explanation_2022} introduce SHAP-based interpretability frameworks for ICU mortality prediction, offering granular insight into model behavior. 
Such frameworks align with our emphasis on validating robustness under uncertainty and suggest directions for future introspective analyses of MMNAR-aware models.

\paragraph{Handling Non-Random Missingness.}
In the context of non-random missingness, \citet{duan_towards_2023} propose PRG, a graph-based semi-supervised framework that improves label quality in MNAR settings, which conceptually parallels our MMNAR formulation. 

\paragraph{Efficient Fusion and Clinical Applications.}
Efficient multimodal fusion has also been studied through architectural innovations. 
\citet{nagrani_attention_2022} propose attention bottlenecks to optimize cross-modal communication while reducing computational costs, highlighting the benefits of architectural constraints for scalability. 
Fusion has also been extended to clinical trial design: AutoTrial~\cite{wang_autotrial_2023} employs discrete-neural prompting to control eligibility criteria generation, illustrating the utility of prompt-based language model interventions in biomedical contexts.

\paragraph{Data Synthesis and Fairness.}
Healthcare applications include prediction under modality incompleteness with M3Care~\cite{zhang2022m3care}, and methods addressing imbalanced data, uncertainty, or fairness (PMSGD~\cite{azad_prediction_2022}, FavMac~\cite{lin_fast_nodate}, FRAMM~\cite{theodorou_framm_2024}). 
In data synthesis, EMIXER~\cite{biswal2022emixer} provides an end-to-end multimodal framework for generating diagnostic image--report pairs, offering strong benefits in low-label regimes. 
These directions highlight how multimodal learning can be extended to fairness, robustness, and decision-support applications beyond clinical prediction.

\section{Extended Experimental Results}
\label{appendix:experiment}

\subsection{Detailed Data Preprocessing}
\label{appendix:data_preprocessing}

\subsubsection{MIMIC-IV Cohort and Inputs}

\paragraph{Data Sources} Structured EHRs, clinical notes, and chest X-rays are linked to the index hospitalization. Modality availability is assessed after applying the observation windows below and reflects both clinical collection and source coverage, including MIMIC-CXR's 2011--2016 coverage of patients with an emergency-department CXR.

\paragraph{Structured Features} For all three MIMIC-IV tasks, diagnostic, laboratory, medication, and other clinical records available by index hospital discharge, including prior health history, are summarized into 2,965 patient-level features. Measurement counts and availability indicators preserve information about clinical data collection.

\paragraph{Text and Imaging Inputs} Discharge summaries and radiology reports describe the index hospitalization under the eligibility rules below. Imaging inputs are restricted to frontal AP/PA views from the latest eligible chest X-ray study, with pretrained image features averaged within that study. Text and image inputs are processed by the supervised modality experts described in Section~\ref{method:preprocess}; their output logits supply the inputs to the fusion projections.

\paragraph{Readmission and ICU Outcomes} Readmission requires a distinct subsequent hospital admission starting within $(0,30]$ days after index discharge. The ICU outcome requires an ICU stay during a subsequent hospital admission, with ICU admission time within $(0,90]$ days after index discharge.

\paragraph{Mortality Outcome Definition} Among adults discharged alive from the index hospitalization, we define 90-day post-discharge all-cause mortality using recorded patient death dates. Death dates combine hospital and state death records \cite{johnson2024mimic}. We normalize the death and index-discharge timestamps to calendar dates and label recorded deaths with a date difference in $[0,90]$ days as positive, including discharge day and day 90. Deaths count regardless of cause or location, including during a subsequent hospitalization. Death dates beyond this window and missing death dates are labeled negative, indicating no recorded death within the outcome window. In-hospital survival determines cohort eligibility rather than the prediction target. Recorded death dates preceding the index discharge are rejected during preprocessing.

\paragraph{Post-discharge Inputs} Readmission, subsequent ICU admission, and mortality share an index-discharge-based retrospective observation protocol. Structured inputs describe the completed index hospitalization, including clinical measurements through discharge, completed hospital and ICU lengths of stay, and discharge disposition. Notes are linked by both patient and admission identifiers. Discharge-summary charted times must fall between admission and 24 hours after discharge; radiology-report charted times must fall between 6 hours before admission and 24 hours after discharge. This documentation grace period allows delayed records, so the information set is retrospective rather than restricted to records available in real time at discharge. CXR acquisition times must fall between admission and discharge. Modality availability indicators use these same eligibility rules. Death dates, outcome labels, and subsequent-admission events are excluded from structured predictors.

\paragraph{Cohort and Training Population} Eligibility requires adulthood, a first recorded hospital admission, an anchor-year group within 2008--2019, and survival to discharge. Excluding nine records with death dates preceding discharge leaves 190,588 eligible patients. The fixed cohort contains 20,000 patients; the remaining 170,588 support training of modality representations. Together with 13,600 development patients, they form a maximum training population of 184,188 before modality-specific eligibility restrictions. The held-out validation and test patients are excluded from this training population.

\subsubsection{eICU Dataset - Virtual Modality Specifications}

We extract information from the patient, diagnosis, treatment, medication, laboratory, and APACHE tables. Eligible ICU stays involve adults aged 18--89 years and last between 12 hours and 10 days. Admissions lacking crucial identifiers are excluded. Variables with more than 80\% missingness are removed, and remaining missing feature values are imputed using the median or zero, as appropriate.

For eICU, structured features are organized into 10 virtual modalities. Availability is determined from the observed feature values before imputation. The clinical groups are defined as follows:
\begin{enumerate}
\item \textbf{Demographics}: Patient characteristics including age, gender, ethnicity, admission weight, height, BMI, and hospital characteristics (teaching status, bed size).
    
\item \textbf{Admission}: Admission source (emergency department, ICU transfer, operating room, etc.) and ICU unit type (MICU, SICU, cardiac ICU, etc.).
    
\item \textbf{APACHE}: Severity-of-illness measures and components including age score, Glasgow Coma Scale components, acute physiology score, predicted ICU and hospital mortality, and ventilation status.
    
\item \textbf{Diagnosis}: Diagnostic information including diagnosis counts, primary diagnostic categories (cardiovascular, respiratory, neurologic, etc.), and surgical patient status.
    
\item \textbf{Medication}: Medication usage patterns including total and unique medication counts, and specific drug categories (antibiotics, vasopressors, sedatives).
    
\item \textbf{Laboratory Values}: Laboratory test results including complete blood count, basic metabolic panel, liver function tests, and statistical aggregations (mean, min, max, standard deviation).
    
\item \textbf{Vital Signs}: Physiological measurements including heart rate, respiratory rate, oxygen saturation, blood pressure (invasive and non-invasive), and derived parameters (shock index, calculated MAP).
    
\item \textbf{Respiratory}: Respiratory support parameters including mechanical ventilation status, fraction of inspired oxygen (FiO$_2$), and positive end-expiratory pressure (PEEP).
    
\item \textbf{Fluid Balance}: Fluid management data including total intake, output, net fluid balance, and positive fluid balance indicators.
    
\item \textbf{Comorbidities}: Pre-existing conditions including AIDS, hepatic failure, malignancies, immunosuppression, and chronic diseases (diabetes, hypertension, heart failure, COPD, chronic kidney disease).
\end{enumerate}

\textbf{Missingness Determination.} The missingness indicator $\delta_i^{(m)} = 1$ if modality $m$ is available for patient $i$ (i.e., $\leq$ 80\% NaN values), and 0 otherwise. This threshold accounts for the clinical reality that some features within a group may be selectively unavailable while maintaining the group's overall clinical utility.

This virtual modality approach enables our MMNAR framework to model systematic patterns in clinical data collection across different hospitals and care protocols, where certain categories of information may be consistently missing due to institutional practices, patient acuity, or resource constraints. The grouped representation allows missingness-aware learning from these structured clinical categories.

\subsection{Training and Validation Protocol}
\label{appendix:protocol}

\paragraph{Supervised Expert Inputs} For MIMIC-IV, the discharge-summary and imaging experts are trained only on eligible patients in the additional training pool outside the fixed cohort. Their fixed predictions supply logits for development, validation, and test patients with the corresponding modality. Structured and radiology experts generate five-fold out-of-fold logits for development patients: each fold's experts are trained on the additional pool and the other four development folds, subject to modality availability. For validation and test prediction, these experts are refitted on the additional pool and all development patients. The structured expert averages logits across three random seeds. The fusion model then learns from the fixed development-patient logits and outcome labels. Validation and test labels are excluded from expert and fusion training. The discharge-summary expert jointly predicts the three main outcomes and auxiliary 90-day readmission, producing its four-dimensional input vector; the other experts supply one logit per main outcome. Modality-wise standardization uses only observed patients in the fusion fitting subset, excluding its internal early-stopping subset.

For MIMIC-IV, the loss weights are $\lambda_{\mathrm{miss}}=0.1$ and $\lambda_{\mathrm{rep}}=0.1$, with equal prediction weights $\lambda_{\mathrm{pred},t}=1$ for all three tasks. The three components are jointly trained through $\mathcal L_{\mathrm{total}}$; pattern-specific rectification follows training.

The 2,400 validation patients are divided into two groups of 1,200 for cross-fitted correction and probability calibration. The threshold $\kappa_t$ and calibration map $C_t$ are fixed using validation data before test outcomes are evaluated. The test set contains 4,000 patients, and no test labels are used for model selection.

\paragraph{Baseline Inputs and Early Stopping} All MIMIC-IV methods use the same eligible population of 190,588 adults with a first recorded hospital admission, survival to discharge, and an anchor-year group within 2008--2019. They share the fixed cohort and patient split, the three outcome labels, and the discharge reference point. Index-hospitalization events are restricted to times no later than discharge, with laboratory measurements eligible from 24 hours before admission. Text uses the same patient/admission matching and charted-time filters described above, including the 24-hour post-discharge documentation window. The shared data snapshot and preprocessing retain the same limitations for every method: magnesium and phosphate laboratory measurements are absent, and measurements originally recorded in Fahrenheit are excluded by the temperature range filter.

The baseline training population is the 170,588 additional patients plus 13,600 development patients, totaling 184,188. All baselines share a fixed internal early-stopping set of 8,529 patients, sampled as 5\% of the additional pool with seed 20261022; the remaining 175,659 patients are used for parameter fitting. This early-stopping set is distinct from both meta partitions. The same overall training population supplies \acronym{}'s experts under the modality-specific restrictions above, while its fusion network uses only development patients and its own internal early-stopping splits.

\paragraph{Baseline Configuration Selection and Calibration} Each baseline declares at most four hyperparameter or architectural configurations before evaluation. We select among them by AP on meta-selection, using AUC as the secondary criterion. Selected baseline predictions are calibrated by logistic regression on their logits, with \texttt{LogisticRegression(C=1e6)} fitted only on meta-calibration. This uses the same Platt calibration family as \acronym{}; the latter's cross-fitted pattern correction and calibration selection follow Section~\ref{method:multitask_outcome}.

\paragraph{Paired Evaluation and Frozen Test Release} All MIMIC-IV methods use AUC, AP computed with \texttt{average\_precision\_score}, and Brier score, together with the same S/I/T/R observation-pattern encoding. Bootstrap indices are generated once with seed 20261022 and shared across methods for 2,000 resamples, so paired metric differences use identical resampled patients. Before scoring, each baseline freezes its code, inputs, model, and generated test predictions with SHA-256 hashes. Baseline label-access routines reject requests for test outcomes during training and selection. Scoring verifies the hashes and writes a persistent scoring lock before reading test labels, permitting a single release per frozen baseline version. All frozen versions are reported, with no selection based on test performance.

\paragraph{Final Ensemble and Confidence Intervals} For MIMIC-IV, we train three fusion models with random seeds 100, 101, and 102 on the development partition, reserving 15\% internally for early stopping in each run. Each task uses the checkpoint with its lowest internal stopping-set binary cross-entropy. We average the three models' task logits with equal weights, apply the sigmoid transformation, and then apply the validation-fitted pattern correction and selected calibration map. The reported \acronym{} point estimates are evaluated on these final ensemble probabilities. Bootstrap evaluation samples 4,000 test patients with replacement in each of 2,000 resamples and uses the 2.5th and 97.5th percentiles of each metric as confidence limits. Models and predictions remain fixed during resampling, so the intervals describe test-patient sampling uncertainty conditional on the fitted ensemble.

\subsection{Performance Across Modality Configurations}
\label{appendix:performance_modality}

Supplementary analyses use the fixed MIMIC-IV cohort and patient split described in Section~\ref{experiment:baselines}. Table~\ref{tab:pattern_outcomes} reports cohort sizes and observed outcome rates for the eight possible modality patterns. Table~\ref{tab:modality_configs} reports pattern-specific predictive performance, while Table~\ref{tab:imputation_comparison} presents the overall comparison with imputation baselines.

\begin{table}[H]
\centering
\small
\setlength{\tabcolsep}{3pt}
\renewcommand{\arraystretch}{1.13}
\caption{Cohort counts and observed outcome rates by modality pattern on MIMIC-IV. S = structured data, I = imaging, T = discharge summaries, and R = radiology reports. Counts describe the fixed cohort ($n=20{,}000$). ICU admission is assessed within 90 days after discharge. Rates are percentages of patients within each pattern. S+I+T is empty, and its event counts and rates are marked by dashes. Rates for S+I and S+I+R are based on very small samples.}
\label{tab:pattern_outcomes}
\fitTable{%
\begin{tabular}{l|r|rr|rr}
\toprule
\textbf{Pattern} & \textbf{Patients} & \multicolumn{2}{c|}{\textbf{30-day Readmission}} & \multicolumn{2}{c}{\textbf{Post-discharge ICU}} \\
& & \textbf{Events} & \textbf{Rate (\%)} & \textbf{Events} & \textbf{Rate (\%)} \\
\midrule
S & 4,734 & 634 & 13.4 & 64 & 1.4 \\
S+R & 2,222 & 193 & 8.7 & 45 & 2.0 \\
S+T & 2,468 & 314 & 12.7 & 65 & 2.6 \\
S+T+R & 9,209 & 1,327 & 14.4 & 422 & 4.6 \\
S+I & 5 & 5 & 100.0 & 0 & 0.0 \\
S+I+R & 69 & 8 & 11.6 & 1 & 1.4 \\
S+I+T & 0 & \textemdash & \textemdash & \textemdash & \textemdash \\
S+I+T+R & 1,293 & 268 & 20.7 & 92 & 7.1 \\
Total & 20,000 & 2,749 & 13.7 & 689 & 3.4 \\
\bottomrule
\end{tabular}
}
\end{table}

\begin{table}[H]
\centering
\small
\setlength{\tabcolsep}{3pt}
\renewcommand{\arraystretch}{1.13}
\caption{Predictive performance across eight possible modality patterns on the MIMIC-IV test set ($n=4{,}000$). Counts refer to test patients. Dashes indicate empty groups or groups with only one outcome class; brackets on the overall row give 95\% confidence intervals.}
\label{tab:modality_configs}
\fitTable{%
\begin{tabular}{l|cc|cc|c}
\toprule
\textbf{Pattern} & \multicolumn{2}{c|}{\textbf{Readmission}} & \multicolumn{2}{c|}{\textbf{ICU Admission}} & \textbf{Test} \\
& \textbf{AUC} & \textbf{AUPRC} & \textbf{AUC} & \textbf{AUPRC} & \textbf{Count} \\
\midrule
S & 0.723 & 0.319 & 0.769 & 0.073 & 928 \\
S+R & 0.820 & 0.406 & 0.885 & 0.327 & 466 \\
S+T & 0.665 & 0.239 & 0.788 & 0.159 & 499 \\
S+T+R & 0.809 & 0.486 & 0.885 & 0.389 & 1,836 \\
S+I & \textemdash & \textemdash & \textemdash & \textemdash & 1 \\
S+I+R & 0.250 & 0.122 & \textemdash & \textemdash & 16 \\
S+I+T & \textemdash & \textemdash & \textemdash & \textemdash & 0 \\
S+I+T+R & 0.822 & 0.568 & 0.895 & 0.535 & 254 \\
Overall (Table~\ref{tab:main_results}) & \shortstack{0.773\\{\scriptsize [0.751, 0.794]}} & \shortstack{0.418\\{\scriptsize [0.374, 0.459]}} & \shortstack{0.868\\{\scriptsize [0.838, 0.896]}} & \shortstack{0.320\\{\scriptsize [0.242, 0.405]}} & 4,000 \\
\bottomrule
\end{tabular}
}
\end{table}

Among the larger incomplete-modality test groups (S, S+R, S+T, and S+T+R), readmission AUC ranges from $0.665$ to $0.820$ and ICU admission AUC from $0.769$ to $0.885$. The complete-modality group has the highest AUC and AUPRC for both tasks. Performance varies with the modality combination and does not increase uniformly with modality count. These comparisons describe different patient subgroups, and AUPRC also reflects their different outcome prevalences. Rare patterns require caution: S+I has only one test patient, S+I+R has 16 patients and a readmission AUC of $0.250$, and S+I+T has no patients. Metrics are omitted for empty groups and groups containing only one outcome class.

\begin{table}[H]
\centering
\small
\setlength{\tabcolsep}{3pt}
\renewcommand{\arraystretch}{1.13}
\caption{Overall comparison with imputation baselines on the MIMIC-IV test set. AUPRC denotes average precision; brackets give 95\% confidence intervals where available.}
\label{tab:imputation_comparison}
\fitTable{%
\begin{tabular}{l|ccc|ccc}
\toprule
\textbf{Method} & \multicolumn{3}{c|}{\textbf{Readmission}} & \multicolumn{3}{c}{\textbf{ICU Admission}} \\
& \textbf{AUC} & \textbf{AUPRC} & \textbf{Brier} & \textbf{AUC} & \textbf{AUPRC} & \textbf{Brier} \\
\midrule
\acronym{} & \shortstack{0.773\\{\scriptsize [0.751, 0.794]}} & \shortstack{0.418\\{\scriptsize [0.374, 0.459]}} & \shortstack{0.0998\\{\scriptsize [0.0931, 0.1066]}} & \shortstack{0.868\\{\scriptsize [0.838, 0.896]}} & \shortstack{0.320\\{\scriptsize [0.242, 0.405]}} & \shortstack{0.0278\\{\scriptsize [0.0235, 0.0324]}} \\
Zero-filling + Masking & 0.706 & 0.310 & 0.1134 & 0.788 & 0.138 & 0.0364 \\
Mean Imputation & 0.680 & 0.280 & 0.1173 & 0.765 & 0.115 & 0.0376 \\
\bottomrule
\end{tabular}
}
\end{table}

In the overall comparison, our framework achieves higher AUC and AUPRC and lower Brier scores than both imputation baselines. Relative to zero-filling with masking, AUC increases by $0.067$ for readmission and $0.080$ for ICU admission. These results support the benefit of our framework for prediction with incomplete multimodal records.

\subsection{Hyperparameter Sensitivity Analysis}
\label{appendix:hyperparameter}

Table~\ref{tab:hyperparam_sensitivity} reports sensitivity to hyperparameter variations for MIMIC-IV readmission, relative to the default AUC $A_0=0.773$ (rounded). Tested settings follow the same training and evaluation protocol as the default configuration. For each parameter, let $A_{\min}$ denote the lowest AUC across the tested settings. We define maximum drop as $D=A_0-A_{\min}$, sensitivity as $100D/A_0$ percent, and stability as $100(1-D/A_0)$ percent. Stability is the complement of sensitivity. Drops and percentages are computed before rounding, so they may not be recovered exactly from the displayed AUCs.

\begin{table}[H]
\centering
\small
\setlength{\tabcolsep}{3pt}
\renewcommand{\arraystretch}{1.13}
\caption{Hyperparameter sensitivity for MIMIC-IV readmission. MMNAR and MR weights are $\lambda_{\mathrm{miss}}$ and $\lambda_{\mathrm{rep}}$. Drops and percentages are computed before rounding; stability is the complement of sensitivity.}
\label{tab:hyperparam_sensitivity}
\fitTable{%
\begin{tabular}{l|c|c|c|c|c|c}
\toprule
\textbf{Parameter} & \textbf{Reference} & \textbf{Parameter Grid} & \textbf{Max Drop} & \textbf{Sensitivity} & \textbf{Stability} & \textbf{AUC Range} \\
\midrule
Learning Rate & 0.001 & \shortstack{$\{0.00031, 0.00067, 0.001,$\\$0.00173, 0.00319\}$} & 0.004 & 0.58\% & 99.42\% & [0.769, 0.773] \\
Weight Decay & 0.0001 & \shortstack{$\{0.000017, 0.000061, 0.0001,$\\$0.000283, 0.000791\}$} & 0.007 & 0.86\% & 99.14\% & [0.766, 0.773] \\
Hidden Dimension & 16 & \shortstack{$\{8, 12, 16,$\\$24, 32\}$} & 0.002 & 0.31\% & 99.69\% & [0.771, 0.773] \\
MMNAR Weight & 0.1 & \shortstack{$\{0.037, 0.081, 0.1,$\\$0.163, 0.247\}$} & 0.004 & 0.49\% & 99.51\% & [0.769, 0.773] \\
MR Weight & 0.1 & \shortstack{$\{0.025, 0.063, 0.1,$\\$0.171, 0.269\}$} & 0.005 & 0.63\% & 99.37\% & [0.768, 0.773] \\
\bottomrule
\end{tabular}
}
\end{table}

Across the tested settings, weight decay produces the largest reported AUC decrease ($0.007$) and relative sensitivity ($0.86\%$). The other parameters have reported relative sensitivities between $0.31\%$ and $0.63\%$, indicating limited sensitivity within the tested ranges for this task.

\subsection{MMNAR Embedding Validation}
\label{appendix:embedding}

To examine the learned missingness embeddings $\bm z_i$, we compare pattern-level norms and observed readmission rates.

\textbf{Systematic Embedding Analysis.} Norms range from $1.508$ to $2.528$ across the seven observed patterns (Table~\ref{tab:embedding_association}). The complete-modality group has the largest norm and a readmission rate of $20.7\%$. The relationship is not monotonic: the S+I group has norm $2.119$ and a readmission rate of $100.0\%$, but contains only five patients. These descriptive comparisons therefore require caution, particularly for rare patterns.

\begin{table}[H]
\centering
\small
\setlength{\tabcolsep}{3pt}
\renewcommand{\arraystretch}{1.13}
\caption{Missingness-embedding norms and observed readmission rates by pattern in the fixed MIMIC-IV cohort. S+I+T is empty; counts are provided in Table~\ref{tab:pattern_outcomes}.}
\label{tab:embedding_association}
\fitTable{%
\begin{tabular}{l|r|r}
\toprule
\textbf{Pattern} & \textbf{Embedding Norm} & \textbf{Readmission Rate (\%)} \\
\midrule
S & 1.888 & 13.4 \\
S+R & 1.508 & 8.7 \\
S+T & 1.820 & 12.7 \\
S+T+R & 2.003 & 14.4 \\
S+I & 2.119 & 100.0 \\
S+I+R & 1.895 & 11.6 \\
S+I+T & \textemdash & \textemdash \\
S+I+T+R & 2.528 & 20.7 \\
\bottomrule
\end{tabular}
}
\end{table}

\textbf{Pattern Coverage.} With S always available and I, T, and R potentially missing, eight patterns are structurally possible. Seven occur in the fixed MIMIC-IV cohort; S+I+T is empty (Table~\ref{tab:pattern_outcomes}). The embedding norms describe variation across these patterns, while the outcome-rate differences provide observational motivation for treating modality availability as informative context.

\subsection{Algorithm}
\label{appendix:algorithm}

Algorithms~\ref{alg:mmnar_fusion}--\ref{alg:multitask_rectifier} summarize the three components. Fusion, reconstruction, and the multitask prediction loss are jointly optimized through $\mathcal L_{\mathrm{total}}$; held-out residual correction follows training.

\LinesNotNumbered

\begin{algorithm2e}[H]
\caption{MMNAR-Aware Modality Fusion (Stage 1A)}
\label{alg:mmnar_fusion}
\KwIn{Observed modalities $\{x_i^{(m)}\}_{m\in\mathcal M_i}$; observation vector $\bm\delta_i$}
\KwOut{Patient representation $\bm h_i$}
Obtain each observed modality embedding $e_i^{(m)}\leftarrow E^{(m)}(x_i^{(m)})$; in MIMIC-IV, compute fixed expert logits $\bm\ell_i^{(m)}\leftarrow F^{(m)}(x_i^{(m)})$, standardize them, and project $e_i^{(m)}\leftarrow P^{(m)}(\tilde{\bm\ell}_i^{(m)})$\;
$\bm z_i\leftarrow\mathrm{MLP}(\bm\delta_i)$\;
Compute the predicted pattern distribution $\hat{\bm\pi}_i$ from $\bm z_i$ and the missingness loss $\mathcal L_{\mathrm{miss}}$\;
\ForEach{$m\in\mathcal M$}{
  $e_{i,\mathrm{gate}}^{(m)}\leftarrow\delta_i^{(m)}\sigma(W^{(m)}\bm z_i+b^{(m)})\odot e_i^{(m)}$, with missing inputs set to zero\;
}
$\bm h_i\leftarrow r_{\mathrm{fuse}}(\{e_{i,\mathrm{gate}}^{(m)}\}_{m\in\mathcal M},\bm\delta_i,\bm z_i)$\;
\Return $\bm h_i$\;
\end{algorithm2e}

\begin{algorithm2e}[H]
\caption{Modality Reconstruction with Latent Alignment (Stage 1B)}
\label{alg:recon_alignment}
\KwIn{Observed modality embeddings and observation vectors; reference embeddings $\bar e_i^{(m)}$}
\KwOut{Reconstruction objective $\mathcal L_{\mathrm{rep}}$}
\ForEach{$m\in\mathcal M$}{
  Randomly select eligible patients $\mathcal A_m$ with $\delta_i^{(m)}=1$ and at least one other observed modality\;
  \If{$|\mathcal A_m|\ge2$}{
    \ForEach{$i\in\mathcal A_m$}{
      Mask modality $m$ and recompute $\bm z_i^{\setminus m}$ and $\bm h_i^{\setminus m}$ using the modified observation pattern\;
      $e_{i,\mathrm{rec}}^{(m)}\leftarrow\phi^{(m)}(\bm h_i^{\setminus m})$\;
    }
    $\mathcal L_{\mathrm{rec}}^{(m)}\leftarrow|\mathcal A_m|^{-1}\sum_{i\in\mathcal A_m}[2-2\,\mathrm{sim}(e_{i,\mathrm{rec}}^{(m)},\operatorname{sg}(\bar e_i^{(m)}))]$\;
  }
  \Else{Set $\mathcal L_{\mathrm{rec}}^{(m)}\leftarrow0$\;}
}
$\mathcal L_{\mathrm{rep}}\leftarrow\lambda_{\mathrm{rep}}\sum_{m\in\mathcal M}\mathcal L_{\mathrm{rec}}^{(m)}$\;
\Return $\mathcal L_{\mathrm{rep}}$\;
\end{algorithm2e}

\begin{algorithm2e}[H]
\caption{Multitask Prediction with Cross-Fitted Rectifier (Stage 2)}
\label{alg:multitask_rectifier}
\KwIn{Shared representations $\bm h_i$; observation patterns $\bm\delta_i$; outcome labels in development and validation data}
\KwOut{Task-specific predictors and final outcome probabilities}
Jointly train task heads with the shared representation using $\mathcal L_{\mathrm{pred}}$\;
Keep the trained encoder and task heads fixed\;
\ForEach{$t\in\mathcal T$}{
  Compute base probabilities $\hat p_{i,t}=g_{\bm\psi'_t}(\bm h_i)$\;
  \For{$k\in\{1,2\}$}{
    Estimate the treatment-effect component $\hat\tau^{(k)}_{\bm\delta,t}$ as the mean residual $y_{i,t}-\hat p_{i,t}$ within each validation-fold pattern, using zero for absent patterns\;
    Apply the thresholded correction to the complementary fold to obtain $q^{\mathrm{cf}}_{i,t}$\;
  }
  Select $\kappa_t$ and the probability calibration map $C_t$ using validation data\;
  Estimate $\hat\tau_{\bm\delta,t}$ from pooled validation residuals\;
  \ForEach{test patient $i$}{
    $q_{i,t}\leftarrow\Pi_\epsilon(\hat p_{i,t}+\mathbf{1}\{|\hat\tau_{\bm\delta_i,t}|>\kappa_t\}\hat\tau_{\bm\delta_i,t})$\;
    $\hat p^{\mathrm{final}}_{i,t}\leftarrow C_t(q_{i,t})$\;
  }
}
\Return $\{g_{\bm\psi'_t},\hat\tau_{\bm\delta,t},\kappa_t,C_t,\hat p^{\mathrm{final}}_{i,t}\}$\;
\end{algorithm2e}

\section{AI Assistants In Research Or Writing}
\label{appendix:ai}
We used AI assistants (e.g., ChatGPT) to support language polishing, LaTeX editing, and phrasing suggestions during the manuscript preparation phase. All conceptual, experimental, and analytical contributions were performed by the authors.

\section{Potential Risks}
\label{appendix:potential}

While our framework demonstrates strong empirical performance, a few potential risks warrant consideration. First, the model relies on several interacting components (e.g., fusion modules, reconstruction objectives, and rectifier corrections), which may increase system complexity and reduce interpretability in clinical deployment scenarios. Second, the framework introduces multiple hyperparameters that may require tuning across different datasets or institutions, potentially limiting out-of-the-box generalizability. Third, although the MMNAR-aware design improves robustness to missing modalities, distributional shifts in missingness patterns across healthcare systems could still affect performance. We leave comprehensive evaluation under such cross-institutional shifts for future work.

\end{document}